\documentclass{article}
\usepackage{spconf,amsmath,graphicx}
\usepackage{amssymb} 
\usepackage{multirow}

\title{3D Unsupervised Region-Aware Registration Transformer}
\name{Yu Hao\textsuperscript{1}, Yi Fang\textsuperscript{1,2}}
\address{\textsuperscript{1}NYU Multimedia and Visual Computing Lab\\
\textsuperscript{2}NYUAD Center for Artificial Intelligence and Robotics\\
New York University Abu Dhabi\\
Abu Dhabi, UAE\\}
\begin{document}
%
\maketitle
\begin{abstract}
This paper concerns the research problem of point cloud registration to find the rigid transformation to optimally align the source point set with the target one. Learning robust point cloud registration models with deep neural networks has emerged as a powerful paradigm, offering promising performance in predicting the global geometric transformation for a pair of point sets. Existing methods first leverage an encoder to regress the global shape descriptor, which is then decoded into a shape-conditioned transformation via concatenation-based conditioning. However, different regions of a 3D shape vary in their geometric structures which makes it more sense that we have a region-conditioned transformation instead of the shape-conditioned one. In this paper, we define our 3D registration function through the introduction of a new design of 3D region partition module that is able to divide the input shape to different regions with a self-supervised 3D shape reconstruction loss without the need for ground truth labels. We further propose the 3D shape transformer module to efficiently and effectively capture short- and long-range geometric dependencies for regions on the 3D shape Consequently, the region-aware decoder module is proposed to predict the transformations for different regions respectively. The global geometric transformation from the source point set to the target one is then formed by the weighted fusion of region-aware transformation. Compared to the state-of-the-art approaches, our experiments show that our 3D-URRT achieves superior registration performance over various benchmark datasets (e.g. ModelNet40).
\end{abstract}
\vspace{-2mm}
\section{Introduction}
\vspace{-2mm}
Point set registration is a challenging but meaningful task, which has wide application in many fields. This task requires us to find the rigid transformation to optimally align the source point set with the target one. 
\begin{figure}
\centering
\includegraphics[width=7cm]{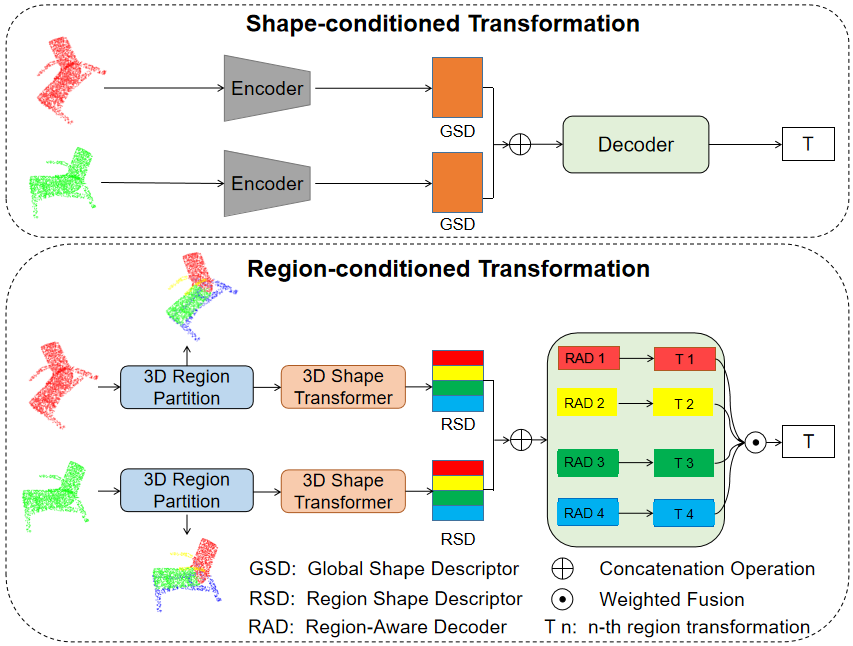}
\vspace{-4mm}
\caption{Comparison between the shape-conditioned transformation and region-conditioned transformation.}
\label{fig:1}
\vspace{-6mm}
\end{figure}
In recent years, deep-learning-based algorithms have been implemented in various industries and achieved great success, researchers are increasingly interested in bringing deep-learning-based solutions to the field of point set registration. These methods usually leverage modern feature extraction technologies for feature learning and then regress the transformation matrix based on the mutual information and correlation defined on the extracted features of source and target shapes. The most representative model, deep closest point (DCP) \cite{wang2019deep}, leverages DGCNN \cite{wang2019dynamic} for feature learning and a pointer network to perform soft matching. To refine the soft matching results to predict the final rigid transformation, the DCP model further proposes a singular value decomposition layer for fine-tuning. However, it is still challenging to design an explicit module for learning both the features from unstructured point clouds and their ``geometric relationship". 
The learning of robust point cloud registration models with deep neural networks \cite{liu2019flownet3d, aoki2019pointnetlk} has emerged as a powerful paradigm, offering promising performance in predicting the global geometric transformation for a pair of point sets. 

As shown in Figure \ref{fig:1}, previous approaches firstly encode the 3D point to a high-dimensional global shape descriptor and use the shape-conditioned decoder to regress the transformation for the given pair 3D shapes. In this paper, we start with our argument that the performance of the previous registration models might be affected by the fact that only the global shape descriptor is used to predict the transformation for a paired 3D shapes. This observation motivate us to develop our proposed 3D Unsupervised Region-Aware Registration Transformer, denoted as 3D-URRT, with the hope to fully utilize both local and global geometric features for more robust 3D shape registration learning. In addition, different regions of a 3D shape vary in their geometric structures which makes it more sense that we have a region-conditioned (in contrast to shape-conditioned) transformation estimation. Note that our proposed 3D-URRT is only trained by a self-supervised 3D shape reconstruction loss and an unsupervised alignment loss without the need of any annotated region labels or transformation ground truth information.

\begin{figure}
\centering
\includegraphics[width=8.5cm]{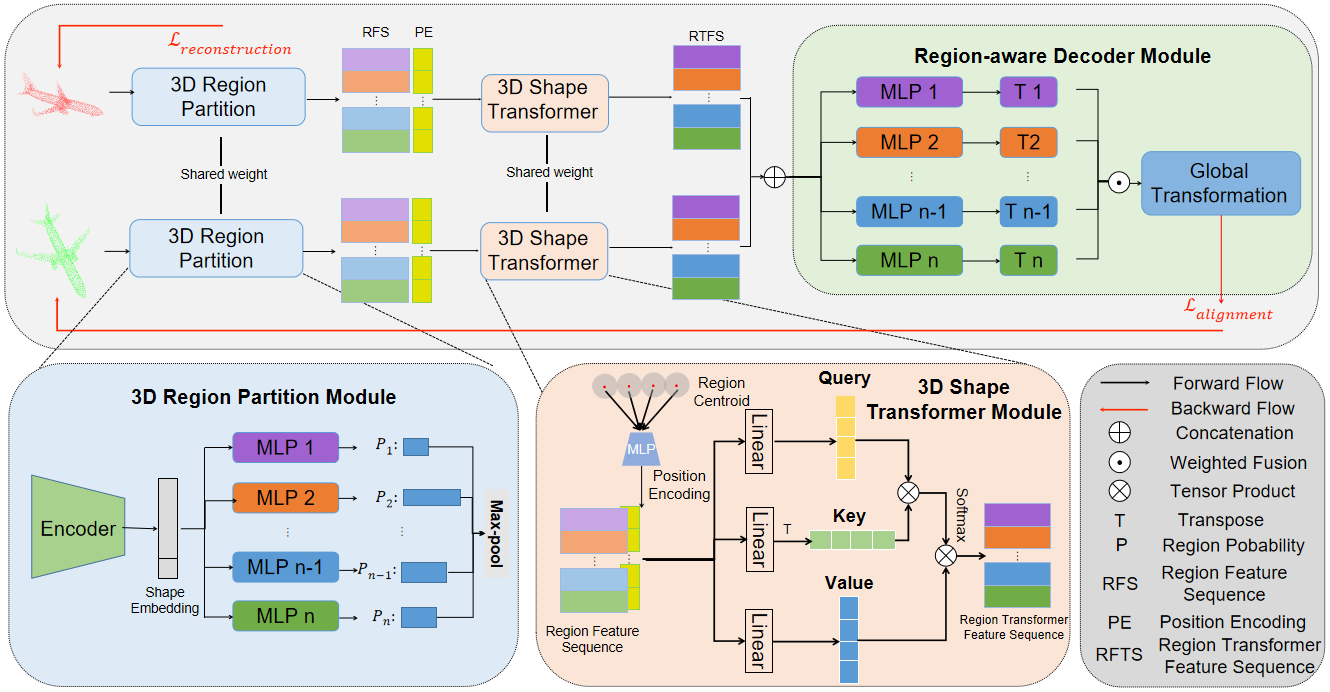}
\vspace{-4mm}
\caption{Our pipeline. Our proposed 3D-URRT framework contains three main components:  3D region partition module, 3D shape transformer module and region-aware decoder module.}
\vspace{-4mm}
\label{main}
\end{figure}

In this paper, as illustrated in Figure \ref{main}, we present 3D region-aware unsupervised registration transformer to predict transformation for pairwise point sets in a self-supervised learning fashion. Our proposed 3D-URRT framework contains three main components. The first component is a 3D region partition module that is responsible to divide the given shape to different 3D regions with a self-supervised 3D shape reconstruction loss without the need for region labels. The second component is the 3D shape transformer module with position encoding that is able to capture short-and long-range geometric dependencies for regions on the 3D shape. The third component is a region-aware decoder module which maps the region-aware transformer features to a set of region-specific transformations. The global geometric transformation from source point set to target one is then formed by weighted fusion of region-aware transformations. Our contribution is as follows: 1) We introduce a new concept of region-conditioned transformation that contributes to a novel 3D region-aware unsupervised registration transformer (3D-URRT) as the learning approach for robust point set alignment. Our 3D-URRT is a novel unsupervised learning model for point cloud registration without the need of training on labeled datasets. 2) We define our 3D registration function through the introduction  of a new design of transformer which is able to efficiently and effectively capture short- and long-range geometric dependencies for regions on the 3D shape. 3) Experimental results demonstrate the effectiveness of the proposed method for point set registration, our 3D-URRT achieved superior performance compared to unsupervised and supervised state-of-the-art approaches even without labeled data for training.
\vspace{-4mm}
\section{Methods}\label{headings}
\vspace{-4mm}

In this section, we first introduce the problem statement of the 3D registration model. Giving a training dataset $D=\{(S, G)\}$, where $S, G \subset \mathbb{R}^3$. $S$ denotes the input source shape and $G$ denotes the input target shape. We aim to obtain a parametric function $g_{\theta}(S, G)$ using a neural network structure that can predict the rotation matrix $ R \in SO(3)$ and the translation vector $t \in \mathbb{R}^3$ that can deform the source point cloud towards the target point cloud. 

\vspace{-2mm}
\subsection{3D Region Partition Module}\label{set1}
\vspace{-2mm}
 In 2D computer vision, by dividing the 2D image into several patches, the transformer is computationally applicable to exploit the relationship between patches and demonstrate excellent performance in different 2D vision tasks. 
 In this paper, we propose the 3D region partition module that is able to divide the 3D shape into different regions with a self-supervised 3D shape reconstruction loss without the need for region labels. Our key idea is to use a set of MLPs to predict probability scores of all parts for each point on the 3D shape. In other words, if two points belong to the same part, they have a stronger response with a higher probability score on the same MLP. With this design, our 3D region partition module first takes as input the concatenation of the point coordinate $x$ and its corresponding shape embedding $e$ extracted by a feature encoder network. Then, the 3D region partition module performs a regression that maps the shape embedding from $\mathbb{R}^{(d + 3)} \to \mathbb{R}^n$ outputting the predict probability scores of all parts for each point, where d denotes the dimension of shape embedding and n denotes the pre-defined number of 3D regions, respectively. Formally, in this model we define a set of non-linear multi-layer perceptron (MLP)-based functions $\{h_k\}_{k=1,...,n}$ with a softmax activation function to predict the probability score $s_k$ that indicates the likelihood that the given point belongs to a particular region: 
\begin{equation}
\begin{split}
s_k= Softmax(h_k([x, e]))
\end{split}
\end{equation}
where $e$ denotes the shape embedding, $x$ denotes a 3D point in the source or target shape and $s_k$ denotes the probability score of the k-th region.

After we have the probability scores of all regions that the given point belongs to, we use a max-pooling function to select the highest probability score among all the pre-defined regions and output the predicted region label $o$ of the input point.
Note that our model is trained with a self-supervised reconstruction loss. The model can output an inside-outside status\cite{chen2019bae} for each point in the shape. While predicting the region label, our model is able to naturally estimate the inside-outside status for each 3D point in the shape. Thus, we can divide the input source or target shape $X$ to several meaningful regions $\{X_k\}_{k=1,2,...,n}$ using the predicted region label $o$, where n denotes the pre-defined number of 3D regions. The region feature sequence (RFS) can be also obtained as $l=\{l_k\}_{k=1,2,...,n}$ by dividing the shape embedding with a max-pooling operation according to the predicted region label.
   
\vspace{-2mm}      
\subsection{3D Transformer Module} \label{set2}
\vspace{-2mm}

Position encoding plays an important role in transformer since it can retain positional information for sequences so that the transformer is able to exploit complex relationships amongst different elements in the sequence \cite{zhao2021point}. In this paper, we propose to use the centroid of the region as the position reference for each region and we utilize a trainable MLP-based position encoding function $\delta: \mathbb{R}^3 \to \mathbb{R}^d$ to encode the centroid coordinate. The final input $f$ of the 3D shape transformer module is the element-wise summation of the encoding feature $p$ and region feature sequence $l$.
To efficiently and effectively capture short- and long-range geometric dependencies for regions on the 3D shape, we define self-attention layers in this section. Given the region feature sequence, self-attention layers are able to estimate the relevance of one region to another region. We can formulate the self-attention layers as:
\begin{equation}
\begin{split}
a_i= \sum_{j=1}^nSoftmax({\varphi(f_i)}^\mathrm{T}\psi(f_j))\alpha(f_i)
\end{split}
\end{equation}
where $\varphi$, $\psi$ and $\alpha$ are linear based feature transformation function $\mathbb{R}^{d} \to \mathbb{R}^d$. The attention weight is the product between features transformed by $\varphi$ and $\psi$. Then we use the Softmax activation function to normalize the attention weight. The region transformer feature sequence (RTFS) defined as $a = \{a_k\}_{k=1,2,...,n}$  (see Figure \ref{main}) is the weighted aggregation between the attention weight and the features. 
\vspace{-2mm}
\subsection{Region-Aware Decoder Module}\label{set3}
\vspace{-2mm}

Different regions of a 3D shape vary in their geometric structures which makes it more sense that we have a region-conditioned (in contrast to shape-conditioned) transformation decoder via concatenation-based conditioning. As shown in Figure \ref{main}, the region-conditioned transformation predicts a set of transformations for different regions, which are then weighted fused to form a global transformation. With this design, we define a set of non-linear MLP-based functions $g_k: \mathbb{R}^{(2m)} \to \mathbb{R}^{c}$, where c is the dimension of output layer. We have the predicted rigid transformation matrix $\phi_k$ as:
$\phi_k= g_k([a^S_k, a^G_k])$
where [,] denotes the operation of concatenation, $\phi_k$ denotes the k-th transformation function in Region-Aware Transformation. $f^S_k$ and $f^G_k$ denote the source point set region feature and the target point set region feature. We use the region point number as the weight to balance among the multiple MLPs from the region-aware decoder module. We define the final transformation matrix $\phi$ as:
\begin{equation}
\begin{split}
\phi= \sum_{k=1}^n\frac{N_k}{N}\phi_k
\end{split}
\end{equation}
Where N is the number of points in source and target shapes and $N_k$ is the number of points in the k-th region of source and target shapes.

\vspace{-2mm}
\subsection{Loss Function}\label{set4}
\vspace{-2mm}
There are two terms in the loss function. We adopt Chamfer Distance, a simple but effective distance metric for alignment loss. The reconstruction loss is defined as a type of inside-outside status \cite{chen2019bae}. In order to obtain the negative points (outside) for training, we sampled a collection of 3D points surrounding the input shape and record the inside-outside status of the sampled points following the sampling method in \cite{chen2019learning}. 


\vspace{-3mm}
\section{Experiments}
\vspace{-2mm}

\subsection{Dataset preparation} \label{exp1}
\vspace{-2mm}
We test the performance of our method for 3D point cloud registration on the ModelNet40 benchmark dataset \cite{wu20153d}. 
For a fair comparision, we follow the exact same experimental setting as DCP\cite{wang2019deep}
For the evaluation of point cloud registration performance, We use the mean squared error (MSE) and mean absolute error (MAE) to measure the performance of our model and all comparing methods. 

\begin{table}
\vspace{-3mm}
  \caption{ Quantitative result. We conduct the ablation study on the ModelNet40 dataset.}
  \label{table1}
  \small
  \centering
  \begin{tabular}{lllllll}
    \hline                  
    Models     &RMSE(R) &MAE(R) &RMSE(t) &MAE(t)\\
    \hline
    Model A &   1.2891 & 1.0016 & 0.0170 & 0.0128    \\
    Model B & 0.6237 & 0.5337 & 0.0097 & 0.0074      \\
    Model C & \textbf{0.5526} & \textbf{0.4328}  & \textbf{0.0080 }& \textbf{0.0060}      \\
    \hline
  \end{tabular}
  \vspace{-4mm}
\end{table}

\begin{table}[t]
\vspace{-2mm}
\caption{ Quantitative result. Comparison using shapes with D.I., P.D. and D.O. noise on the ModelNet40 dataset.}
\label{table2}
\small
\centering
\resizebox{\linewidth}{!}{
\begin{tabular}{l|l|llll}
\hline
Noise                       & Models & RMSE(R) & MAE(R) & RMSE(t) & MAE(t) \\ \hline
\multirow{3}{*}{D.I.} & PR-NET & 3.9304  & 2.9692 & 0.0229  & 0.0175 \\
                            & DCP    & 5.6053  & 4.4149 & 0.0310  & 0.0234 \\
                            & Ours   & \textbf{2.2623}  & \textbf{1.6740} & \textbf{0.0187}  & \textbf{0.0133} \\ \hline
\multirow{3}{*}{P.D.} & PR-NET & 2.7431  & 2.1505 & 0.0224  & 0.0177 \\
                            & DCP    & 3.9905  & 3.0104 & 0.0220  & 0.0170 \\
                            & Ours   & \textbf{1.2238}  & \textbf{0.9811} & \textbf{0.0152}  & \textbf{0.0115} \\ \hline
\multirow{3}{*}{D.O.} & PR-NET & 4.6381  & 3.5596 & 0.0247  & 0.0189 \\
                            & DCP    & 7.1032  & 5.7792 & 0.0349  & 0.0271 \\
                            & Ours   & \textbf{2.3787}  & \textbf{1.6398} & \textbf{0.0217}  & \textbf{0.0169} \\ \hline
\end{tabular}}
\vspace{-3mm}
\end{table}

\begin{figure}[t]
\centering
\includegraphics[width=7.5cm]{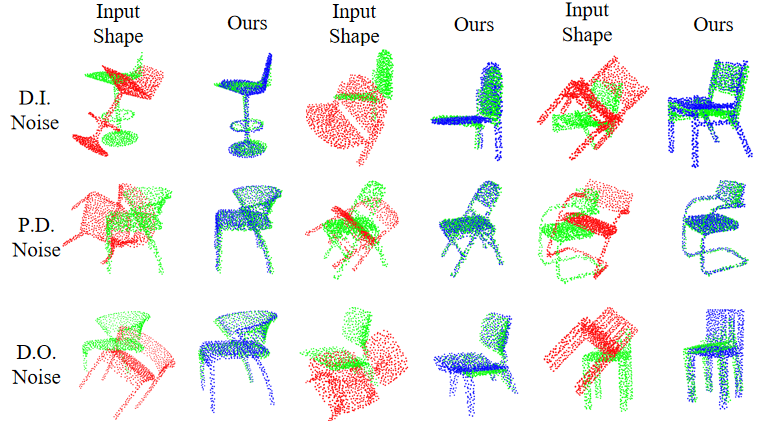}
\vspace{-3mm}
\caption{ Qualitative results. Randomly selected qualitative results in presence of D.I., P.D., and D.O. noises on the ModelNet40 dataset.}
\vspace{-5mm}
\label{fig:4}
\end{figure}

\begin{table} [t] 
\vspace{-2mm}
  \caption{ Quantitative result. Comparison with SOTA using shapes on the ModelNet40 dataset.}
  \label{table5}
  \centering
  \small
  \begin{tabular}{lllll}
    \hline                  
    Models     & MSE(R)  &MAE(R) &MSE(t)  &MAE(t)\\
    \hline
    ICP \cite{besl1992method}& 894.8973  &23.5448 &0.0846  &0.2487    \\
    PNLK \cite{aoki2019pointnetlk}& 227.8703  &4.2253 &0.0004  &0.0054    \\
    GO-ICP \cite{yang2015go} & 140.4773  & 2.5884 & 0.0006  & 0.0070     \\
    FGR \cite{zhou2016fast} & 87.6614  &1.9992 &0.0001  &0.0028
     \\
    DGMR \cite{yuan2020deepgmr} &  7.9106  &2.8125 &0.0001  &0.0100     \\
    DCPv1 \cite{wang2019deep} & 6.4805  &1.5055 &0.000003  &0.001451     \\
    RPM-Net \cite{yew2020rpm} &  4.7284  &1.3847 &0.000078  &0.005367     \\
    DCPv2 \cite{wang2019deep} &  1.3073 &0.7705 &0.000003  &0.001195      \\
    \hline
    Ours & \textbf{0.3223}  & \textbf{0.4328} & 0.000070  & 0.006012      \\
    \hline
  \end{tabular}
  \vspace{-5mm}
\end{table}

\vspace{-3mm}
\subsection{Ablation study} \label{exp3}
\vspace{-2mm}

\noindent{\textbf{Experiment setting:}} 
For the experimental setting of model A, without our regional-aware decoder, we use the classical shape-conditioned decoder (one decoder) for point cloud registration. For the experimental setting of model B, we utilize the 3D region partition module and 3D shape transformer module for region partition and region-aware feature learning. The region-aware decoder is also leveraged to estimate the region-aware transformation for each region. The global geometric transformation is then formed by the weighted fusion of region-aware transformation. For the experimental setting of model C, we further integrate the position encoding feature as the additional input to 3D shape transformer module. 

\noindent{\textbf{Results:} }
By comparing results of model A in Table \ref{table1}, the performance gain of model B indicates the effectiveness of 3D shape transformer module as a technique to capture short- and long-range geometric dependencies for region-aware shape descriptor learning. This results further explain why it is more sense we have region-conditioned transformation instead of shape-conditioned one. By comparing the performance of model B and model C, we validate the effectiveness of the proposed position encoding. We observed a good performance improvement that indicates our proposed position encoding is able to retain the relative position information of each 3D region. 
 
\vspace{-5mm}
\subsection{Studies on Resistance to Noise} \label{exp4}
\vspace{-2mm}
\noindent{\textbf{Experiment setting:} }
In this experiment, we conduct the experiments to verify our model’s performance using Data Incompleteness (D.I.) Noise, Point Drifts (P.D.) Noise and Data Outliers (D.O.) Noise on 3D shapes. As for D.I. noise (partial), we randomly select a point in unit space and keep its 768 nearest neighbor points. As for P.D. noise, we randomly add Gaussian noise on the entire shapes, which is randomly sampled from N (0, 0.01) and clipped to [-0.05, 0.05]. As for D.O. noise, we first remove a certain amount of points and randomly add the same amount of points generated by a zero-mean Gaussian to the entire point clouds. 

\noindent{\textbf{Results:} }
We list the quantitative experimental results about comparison using shapes with several noises in Table \ref{table2}. The table presents that our method achieves remarkably better performance than PR-NET and DCP models regarding the translation prediction and rotation angle results on the ModelNet40 dataset. In addition, from the qualitative results shown in Figure \ref{fig:4}, we notice that our model achieves remarkable alignment result with D.I. noise, P.D. noise and D.O. noise. 






\vspace{-3mm}
\subsection{Comparisons with state-of-the-art methods} \label{exp7}
\vspace{-2mm}
\noindent{\textbf{Experiment setting:}} In this experiment, we evaluate the
overall registration performance of our proposed model on multiple shape categories and yield the performance compared to well applied and current state-of-the-art methods. For a fair comparison, following exactly DCP’s setting, we split the dataset into 9,843 models for training and 2,468 models for testing. Note that our model is trained without using any ground-truth information and our model does not require the SVD-based fine-tuning processes.                  

\noindent{\textbf{Results:} }
We list the quantitative experimental results in Table \ref{table5}. The data demonstrate that our model, as an unsupervised method, possesses excellent generalization ability. Even though our approach does not require label information for training purposes and an additional SVD layer for fine-tuning, our model still has significantly better performances than DCPv2 (supervised) version. Also, our model is more robust to random point sampling of source and target shapes by adopting the Chamfer Distance loss, whereas the DCP would have severe degradation since it by default assumes the same sampling of points. One may note that the translation vector prediction performance of our model is inferior to that of DCPv1, DCPv2. The reason for this gap is that DCPv2 adopts an additional attention mechanism in its network for enhancement. DCPv1/DCPv2 leverage SVD as an additional fine-tuning the process to refine their results. Compared to other unsupervised algorithms, like ICP and FGR, the strength and accuracy of our model could be clearly observed.

\vspace{-5mm}
\section{Conclusion}
\vspace{-3mm}
In this paper, we present 3D region-aware unsupervised registration transformer, denoted as 3D-URRT, to predict transformation for pairwise point sets in the self-supervised learning fashion. Compared with previous shape-conditioned transformation methods, with the proposed region-aware transformation network, our model can learn the desired geometric transformations for multiple regions in a particular shape which makes the model has great generalization ability and more robust to the noises. In addition, we propose the 3D shape transformer module which is able to efficiently and effectively capture short- and long-range geometric dependencies for regions on the 3D shape. Our proposed method is trained in an unsupervised manner without any ground-truth labels. 


\vspace{-5mm}
\section{ACKNOWLEDGEMENT}
\vspace{-3mm}
This work was supported by the NYUAD Center for
Artificial Intelligence and Robotics (CAIR), funded by Tamkeen under the NYUAD
Research Institute Award CG010.

\bibliographystyle{IEEEbib}
\bibliography{egbib}

\begin{thebibliography}{10}

\bibitem{wang2019deep}
Yue Wang and Justin~M Solomon,
\newblock ``Deep closest point: Learning representations for point cloud
  registration,''
\newblock in {\em Proceedings of the IEEE International Conference on Computer
  Vision}, 2019, pp. 3523--3532.

\bibitem{wang2019dynamic}
Yue Wang, Yongbin Sun, Ziwei Liu, Sanjay~E Sarma, Michael~M Bronstein, and
  Justin~M Solomon,
\newblock ``Dynamic graph cnn for learning on point clouds,''
\newblock {\em ACM Transactions on Graphics (TOG)}, vol. 38, no. 5, pp. 1--12,
  2019.

\bibitem{liu2019flownet3d}
Xingyu Liu, Charles~R Qi, and Leonidas~J Guibas,
\newblock ``Flownet3d: Learning scene flow in 3d point clouds,''
\newblock in {\em Proceedings of the IEEE Conference on Computer Vision and
  Pattern Recognition}, 2019, pp. 529--537.

\bibitem{aoki2019pointnetlk}
Yasuhiro Aoki, Hunter Goforth, Rangaprasad~Arun Srivatsan, and Simon Lucey,
\newblock ``Pointnetlk: Robust \& efficient point cloud registration using
  pointnet,''
\newblock in {\em Proceedings of the IEEE Conference on Computer Vision and
  Pattern Recognition}, 2019, pp. 7163--7172.

\bibitem{chen2019bae}
Zhiqin Chen, Kangxue Yin, Matthew Fisher, Siddhartha Chaudhuri, and Hao Zhang,
\newblock ``Bae-net: Branched autoencoder for shape co-segmentation,''
\newblock in {\em Proceedings of the IEEE International Conference on Computer
  Vision}, 2019, pp. 8490--8499.

\bibitem{zhao2021point}
Hengshuang Zhao, Li~Jiang, Jiaya Jia, Philip~HS Torr, and Vladlen Koltun,
\newblock ``Point transformer,''
\newblock in {\em Proceedings of the IEEE/CVF International Conference on
  Computer Vision}, 2021, pp. 16259--16268.

\bibitem{chen2019learning}
Zhiqin Chen and Hao Zhang,
\newblock ``Learning implicit fields for generative shape modeling,''
\newblock in {\em Proceedings of the IEEE/CVF Conference on Computer Vision and
  Pattern Recognition}, 2019, pp. 5939--5948.

\bibitem{wu20153d}
Zhirong Wu, Shuran Song, Aditya Khosla, Fisher Yu, Linguang Zhang, Xiaoou Tang,
  and Jianxiong Xiao,
\newblock ``3d shapenets: A deep representation for volumetric shapes,''
\newblock in {\em Proceedings of the IEEE Conference on Computer Vision and
  Pattern Recognition}, 2015, pp. 1912--1920.

\bibitem{besl1992method}
Paul~J Besl and Neil~D McKay,
\newblock ``Method for registration of 3-d shapes,''
\newblock in {\em Sensor Fusion IV: Control Paradigms and Data Structures}.
  International Society for Optics and Photonics, 1992, vol. 1611, pp.
  586--607.

\bibitem{yang2015go}
Jiaolong Yang, Hongdong Li, Dylan Campbell, and Yunde Jia,
\newblock ``Go-icp: A globally optimal solution to 3d icp point-set
  registration,''
\newblock {\em IEEE transactions on pattern analysis and machine intelligence},
  vol. 38, no. 11, pp. 2241--2254, 2015.

\bibitem{zhou2016fast}
Qian-Yi Zhou, Jaesik Park, and Vladlen Koltun,
\newblock ``Fast global registration,''
\newblock in {\em European Conference on Computer Vision}. Springer, 2016, pp.
  766--782.

\bibitem{yuan2020deepgmr}
Wentao Yuan, Benjamin Eckart, Kihwan Kim, Varun Jampani, Dieter Fox, and Jan
  Kautz,
\newblock ``Deepgmr: Learning latent gaussian mixture models for
  registration,''
\newblock in {\em European Conference on Computer Vision}. Springer, 2020, pp.
  733--750.

\bibitem{yew2020rpm}
Zi~Jian Yew and Gim~Hee Lee,
\newblock ``Rpm-net: Robust point matching using learned features,''
\newblock in {\em Proceedings of the IEEE/CVF conference on computer vision and
  pattern recognition}, 2020, pp. 11824--11833.

\end{thebibliography}

\end{document}